\documentclass[letterpaper, 10 pt, conference]{ieeeconf}  % Comment this line out if you need a4paper

\IEEEoverridecommandlockouts                              % This command is only needed if 
\usepackage{graphics} % for pdf, bitmapped graphics files
\usepackage{epsfig} % for postscript graphics files
\usepackage{mathptmx} % assumes new font selection scheme installed
\usepackage{times} % assumes new font selection scheme installed
\usepackage{amsmath} % assumes amsmath package installed
\usepackage{amssymb}  % assumes amsmath package installed

\usepackage[ruled, longend, linesnumbered, vlined]{algorithm2e}
\usepackage{setspace}
\usepackage{multirow}
\usepackage{subcaption}
\usepackage{url}
\usepackage{booktabs} 
\usepackage{caption}
\usepackage{tabularx}
\usepackage{xcolor}
\usepackage{adjustbox}
\usepackage{pifont}% http://ctan.org/pkg/pifont
\usepackage{tikz}
\usepackage[table]{xcolor}
\definecolor{best}{RGB}{222,235,247}
\definecolor{best_grey}{RGB}{230,230,230}

\definecolor{tblGreen}{HTML}{00A087}
\definecolor{tblBlue}{HTML}{4DBBD5}
\definecolor{tblRed}{HTML}{E64B35}
\definecolor{tblGray}{HTML}{8A8A8A}
\newcommand{\cmark}{\textcolor{tblGreen}{\ding{51}}}%
\newcommand{\xmark}{\textcolor{tblRed}{\ding{55}}}%
\newcommand{\namark}{\textcolor{tblGray}{n/a}}%
\newcommand{\bestcell}[1]{\cellcolor{tblGreen!18}\textbf{#1}}
\newcommand{\secondcell}[1]{\cellcolor{tblBlue!12}#1}

\newcommand{\delayvalue}[2]{\tikz[remember picture, baseline=(#1.base)] \node[inner sep=0pt] (#1) {#2};}
\newcommand{\deltaspace}{\makebox[4.4em][c]{\strut}}
\newcommand{\drawdelayarrow}[3]{%
  \draw[->, >=stealth, draw=black!78, line width=0.45pt]
    (#1.east) .. controls +(1.25em,0) and +(1.25em,0) ..
    node[pos=0.52, anchor=west, xshift=0.5em, inner sep=0pt, font=\scriptsize] {#3}
    (#2.east);%
}

\title{\LARGE \bf
CoRL-MPPI: Enhancing MPPI With Learnable Behaviours For Efficient And Provably-Safe Multi-Robot Collision Avoidance
}

\author{Stepan Dergachev, Artem Pshenitsyn, Aleksandr Panov, Alexey Skrynnik, Konstantin Yakovlev}%

\begin{document}

\maketitle
\thispagestyle{empty}
\pagestyle{empty}

%%%%%%%%%%%%%%%%%%%%%%%%%%%%%%%%%%%%%%%%%%%%%%%%%%%%%%%%%%%%%%%%%%%%%%%%%%%%%%%%
\begin{abstract}

Decentralized collision avoidance is a core challenge for scalable multi-robot systems. A promising approach to this problem is Model Predictive Path Integral (MPPI) control -- a framework that naturally handles arbitrary motion models and provides strong theoretical guarantees. Still, in practice an MPPI-based controller may produce suboptimal trajectories because its performance relies heavily on uninformed random sampling. We introduce CoRL-MPPI, a fusion of Cooperative Reinforcement Learning and MPPI that addresses this limitation. We train an action policy, approximated by a deep neural network, in simulation to learn local cooperative collision-avoidance behaviors. This learned policy is then embedded into the MPPI framework to guide its sampling distribution, biasing it toward more intelligent and cooperative actions in scenarios that may differ substantially from those used during training. Moreover, CoRL-MPPI preserves the theoretical guarantees of regular MPPI. We evaluate our approach in dense, dynamic setups against classical and learning-based state-of-the-art baselines. Our results demonstrate that CoRL-MPPI outperforms competing methods and significantly improves navigation efficiency, measured by success rate and delay, as well as safety, enabling agile and robust multi-robot navigation.

\end{abstract}

%%%%%%%%%%%%%%%%%%%%%%%%%%%%%%%%%%%%%%%%%%%%%%%%%%%%%%%%%%%%%%%%%%%%%%%%%%%%%%%%
\section{INTRODUCTION}

The deployment of multi-robot systems promises a significant boost in efficiency in warehouse logistics, search-and-rescue, disaster management, etc. A fundamental problem in a multi-robot system is decentralized collision avoidance: each robot must navigate to its goal while proactively avoiding conflicts with the others. This problem is inherently challenging due to the non-linear nature of robot interactions, the curse of dimensionality as the number of agents grows, and the necessity for real-time computation under uncertainty.

The known approaches to this problem are mostly reactive. Typical examples include Velocity Obstacles and Optimal Reciprocal Collision Avoidance (ORCA)~\cite{van2011reciprocal_n}, which compute collision-free velocities based on the current states of neighboring robots. 
While highly computationally efficient, reactive methods are inherently myopic. They operate on a one-step time horizon, which can lead to oscillatory behavior, deadlocks, and a general lack of cooperation.

Conversely, methods based on receding-horizon optimal control, most notably the Model Predictive Control (MPC) framework, explicitly optimize a trajectory over multiple steps while accounting for predicted future states. The Model Predictive Path Integral (MPPI)~\cite{williams2016aggressive}, a sampling-based variant of MPC, has gained significant attention for its ability to handle non-linear dynamics and complex cost functions without the need for gradient computation. MPPI allows flexible formulation of both motion models and cost functions and is widely used in mobile robotics.

Still, the performance of MPPI-based methods is critically dependent on the quality of its sampled trajectories. In its standard formulation, control sequences are drawn from a Gaussian distribution centered around a prior (often the previous solution). Such sampling may be very inefficient in complex multi-agent settings, where the vast majority of sampled trajectories may lead to uncooperative behavior. Consequently, even when the number of samples is high, the resultant trajectories may be overly egoistic, leading to an overall degradation of the multi-robot system's performance. Generally, one may claim that \textit{MPPI in multi-robot navigation lacks an understanding of multi-agent cooperation}.

\begin{figure}[t!]
	\centering
	\includegraphics[width=0.98\linewidth]{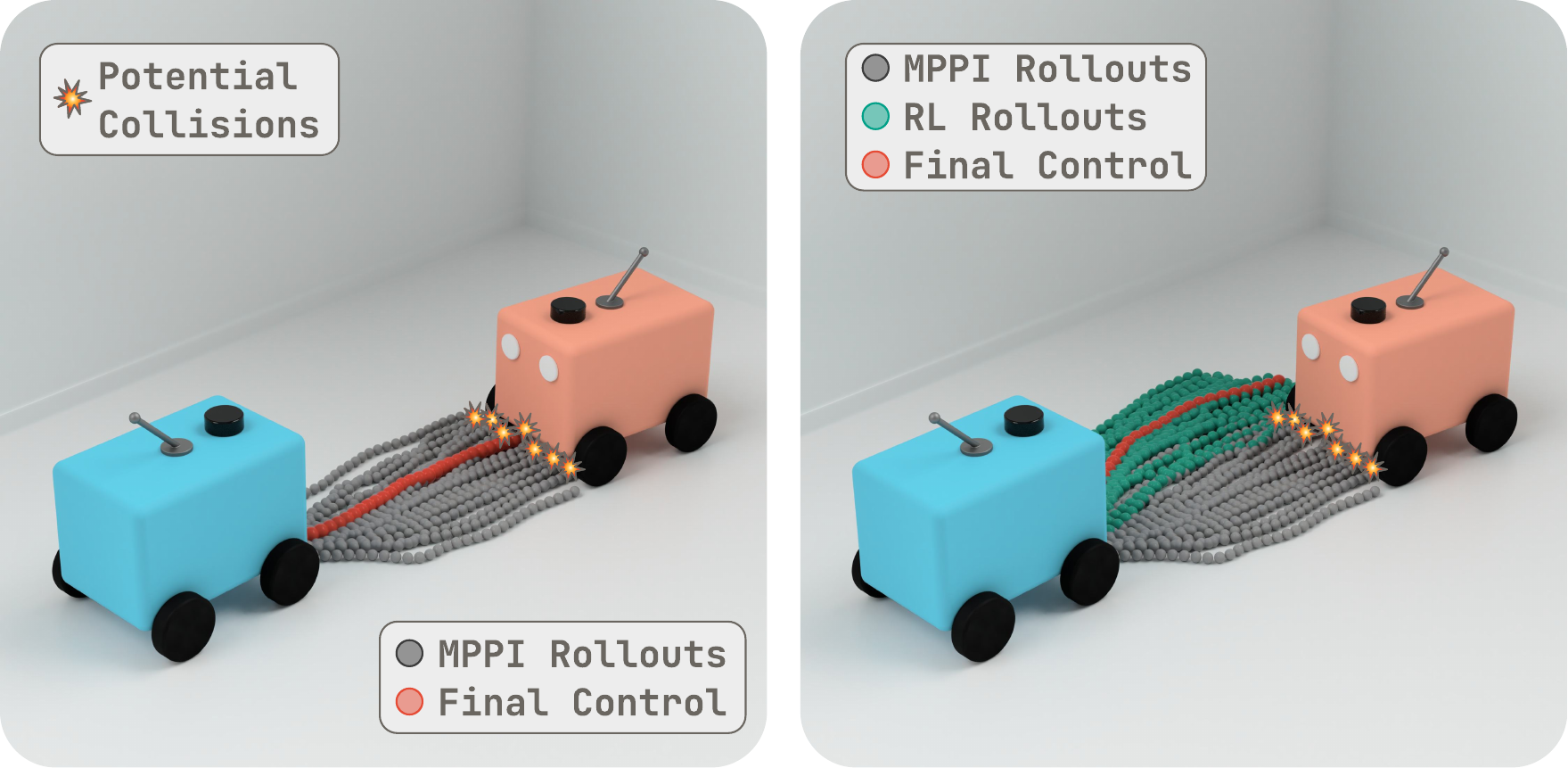}
	\caption{Left: the standard MPPI controller samples random rollouts (grey), which can lead to collisions and suboptimal control (red trajectory). Right: the proposed fusion of RL and MPPI uses learned policy rollouts (blue) to bias sampling toward cooperative, collision-free behavior.}
	\label{fig:visual_abstract}
\end{figure}

To this end, we first train a behavior policy in simulation via reinforcement learning that encapsulates the high-level strategic knowledge of cooperative avoidance and then use the distribution of controls from this policy for sampling within MPPI. Thus the search is biased towards intelligent and cooperative trajectories (see Fig.~\ref{fig:visual_abstract}) while retaining all the theoretical guarantees. The resultant method is called CoRL-MPPI. We believe we are the first to develop such an approach in (i) fully decentralized, (ii) partially observable, and (iii) multi-agent setting.

To summarize, the core contributions of this paper are:
\begin{itemize}
    \item We introduce CoRL-MPPI, a novel hybrid architecture that integrates a learned RL policy into the MPPI control framework to guide its sampling distribution for decentralized multi-robot navigation.
    \item We show theoretically that CoRL-MPPI preserves safety guarantees, even under execution noise.
    \item We conduct comprehensive empirical evaluation in simulation demonstrating that CoRL-MPPI outperforms both classical and learnable state-of-the-art collision avoidance algorithms.
    \item We further validate \textsc{CoRL-MPPI} in a physics-based Gazebo simulation, demonstrating collision-free navigation and real-time performance in a ROS-based pipeline.
\end{itemize}

\section{RELATED WORKS}

\begin{table}[htb!]
\centering
\small
\renewcommand{\arraystretch}{1.1}
\begin{tabular}{lcccccc}
\toprule

Method &
\adjustbox{angle=75,lap=\width-(0.4cm)}{\shortstack{Multi-agent}} &
\adjustbox{angle=75,lap=\width-(0.4cm)}{\shortstack{Decentralized}} &
\adjustbox{angle=75,lap=\width-(0.4cm)}{\shortstack{Cooperative}} &
\adjustbox{angle=75,lap=\width-(0.4cm)}{\shortstack{Safety\\guarantees}} &
\adjustbox{angle=75,lap=\width-(0.4cm)}{\shortstack{Kinematic\\constraints}} &
\adjustbox{angle=75,lap=\width-(0.4cm)}{\shortstack{Execution\\noise}} \\

\midrule
\textsc{ORCA}~\cite{snape2010smooth, alonso2013optimal}
& \cmark
& \cmark
& \xmark
& \cmark
& \cmark$^*$
& \xmark\\

\textsc{BVC}~\cite{zhou2017fast, zhu2022decentralized}
& \cmark
& \cmark
& \xmark
& \cmark
& \cmark$^*$
& \xmark \\

\textsc{D4ORM}~\cite{zhang2025d4orm}
& \cmark
& \xmark
& \cmark
& \xmark
& \cmark 
& \xmark\\

\textsc{RL-RVO}~\cite{han2022reinforcement}
& \cmark
& \cmark
& \cmark
& \xmark
& \cmark$^*$
& \xmark\\

\textsc{RL-{DRIVEN} MPPI}~\cite{qu2024rldriven}
& \xmark
& \namark
& \namark
& \xmark
& \cmark 
& \cmark \\

\textsc{MPPI-ORCA}~\cite{dergachev2025decentralized}
& \cmark
& \cmark
& \xmark
& \cmark
& \cmark
& \cmark \\

\midrule
\textsc{CoRL-MPPI} (ours)
& \cmark
& \cmark
& \cmark
& \cmark
& \cmark 
& \cmark \\
\bottomrule
\end{tabular}
\caption{Comparison with approaches closely related to CoRL-MPPI in multi-agent collision avoidance and RL-enhanced predictive control. \cmark$^{*}$ indicates limited kinematic support: the method is restricted to predefined motion models or requires additional model-specific components for new robot kinematics (e.g. dedicated controller).}
\label{tab:comparison}
\end{table}

This section reviews representative approaches related to our work and highlights the key limitations that motivate the proposed method.

A large family of multi-agent collision avoidance methods is based on geometric formulations such as Velocity Obstacles~\cite{van2011reciprocal_n, snape2010smooth, alonso2013optimal} and Buffered Voronoi Cells~\cite{zhou2017fast, zhu2022decentralized}. Representative examples include ORCA-DD~\cite{snape2010smooth}, NH-ORCA~\cite{alonso2013optimal} and B-UAVC~\cite{zhu2022decentralized}. These approaches provide decentralized collision avoidance and can account for kinematic constraints. However, their support for robot kinematics is often limited to specific motion models or requires additional model-dependent components, such as lookup tables and/or dedicated controllers. Furthermore, they do not explicitly optimize cooperative behavior and may become inefficient in dense interaction scenarios.

Learning-based methods~\cite{chen2017decentralized,long2018towards,fan2020distributed,han2022reinforcement} attempt to improve coordination by learning navigation policies directly from data. For example, RL-RVO~\cite{han2022reinforcement} combines reinforcement learning with reciprocal velocity obstacles to generate decentralized collision-avoidance behaviors. While such methods often exhibit cooperative behavior, they generally lack formal safety guarantees and may struggle to generalize beyond the scenarios encountered during training.

Another important direction is sampling-based planning and control, which can naturally handle nonlinear dynamics, complex cost functions, and long-horizon interactions. Model Predictive Path Integral (MPPI) control~\cite{williams2016aggressive, williams2017information} has become one of the most successful frameworks in this area and has inspired numerous extensions focused on robustness and constraint handling~\cite{gandhi2021robust, yin2022trajectory, tao2022control}. More recently, sampling-based methods have also been applied to multi-agent navigation. For example, D4ORM~\cite{zhang2025d4orm} generates collision-free kinodynamically feasible trajectories for multiple robots through centralized optimization, while MPPI-ORCA~\cite{dergachev2024model,dergachev2025decentralized} provides decentralized collision avoidance with formal safety guarantees. However, D4ORM relies on centralized planning and assumes accurate trajectory tracking, whereas MPPI-ORCA does not explicitly encourage cooperative behavior.

Recent works have also explored the integration of reinforcement learning and sampling-based predictive control. TD-MPC~\cite{hansen2024td} and RL-driven MPPI~\cite{qu2024rldriven} leverage learned policies and value functions to improve sample efficiency, but remain limited to single-agent settings and do not provide formal safety guarantees.

As summarized in Table~\ref{tab:comparison}, existing approaches typically satisfy only a subset of the desired properties. In contrast, the proposed CoRL-MPPI combines decentralized operation, explicit cooperation, formal safety guarantees, consideration of kinematic constraints, and robustness to execution uncertainty within a single framework.

\section{PROBLEM STATEMENT} \label{sec:problem_statement}

Consider a set of homogeneous robots (agents) $\mathcal{A} = \{1, 2, \dots, N\}$, operating in a two-dimensional workspace $\mathcal{W} \subset \mathbb{R}^2$. Each robot is modeled as a disk of radius $r$. Time is discretized, and at each step, every robot selects a control (action) $\mathbf{u}_t \in \mathbb{R}^m$ to update its state $\mathbf{x}_t \in \mathbb{R}^n$. The executed control is subject to stochastic perturbations that model actuation uncertainty:
\begin{equation}
    \mathbf{\nu}_t \sim \mathcal{N}(\mathbf{u}_t, \Sigma), \quad \Sigma = \operatorname{diag}(\sigma_1^2, \dots, \sigma_m^2)
\end{equation}
where $\mathbf{\nu}_t \in \mathbb{R}^m$ represents the actual, randomly perturbed control signal.

The robot kinematic model is described in a general form and the control input is bounded:
\begin{equation}
% \label{eq:aff_dyn}
\label{eq:control_bounds}
\begin{aligned}
    \mathbf{x}_{t+1} = F(\mathbf{x}_t) + G(\mathbf{x}_t)\mathbf{\nu}_t,\\
    \mathbf{\nu}_{\min}[k] \leq \mathbf{\nu}_t[k] \leq \mathbf{\nu}_{\max}[k]
\end{aligned}
\end{equation}
where $F : \mathbb{R}^n \to \mathbb{R}^n$ and $G : \mathbb{R}^n \to \mathbb{R}^{n \times m}$ are given functions (capturing robot's kinematic constraints), $[k]$ denotes the $k$-th element of a vector.

At each time step, robot $i$ has perfect knowledge of its own state~$\mathbf{x}^i_t$
and can perceive the relative positions $\mathbf{p}^j_t$ and velocities $\mathbf{v}^j_t$ of the nearby robots within a certain range.

From a learning perspective, this decentralized navigation problem can be formulated as a decentralized partially observable Markov decision process (Dec-POMDP)~\cite{bernstein2002complexity}. We write it as a tuple $(\mathbb{X}, \{\mathbb{U}^i\}_{i=1}^{N}, \mathbb{T}, \Re, \{\Omega^i\}_{i=1}^{N}, \gamma)$, where $\mathbb{X}$ is the global state space, $\mathbb{U}^i$ is the action space of robot $i$, $\mathbb{T}(\mathbf{x}_{t+1}^{\times}|\mathbf{x}_t^{\times}, \mathbf{u}_t^{\times})$ is the transition model induced by~\eqref{eq:control_bounds}, $\Re$ is the reward function, $\Omega^i:\mathbb{X}\rightarrow\mathbb{O}^i$ maps the global state to the local observation $\mathbf{o}_t^i$ of robot $i$, and $\gamma \in [0,1)$ is the discount factor. A decentralized policy $\pi^i(\mathbf{o}_t^i)$ selects actions using only local information, and the joint policy $\pi^{\times}=\prod_{i=1}^{N}\pi^i$ is trained to maximize the expected discounted return:
\begin{equation}
\label{eq:dec_pomdp_objective}
    \max_{\pi^1,\ldots,\pi^N}
    \mathbb{E}_{\pi^{\times},\mathbb{T}}
    \left[
    \sum_{t=0}^{\infty}
    \gamma^t \Re(\mathbf{x}^{\times}_t, \mathbf{u}^{\times}_t, \mathbf{x}^{\times}_{t+1})
    \right].
\end{equation}
This formulation is used only to train the cooperative proposal policy; at execution time each robot still solves the safe control problem below.

The control $\mathbf{u}^i_t$ of robot $i$ is said to be \emph{probabilistically safe} (or simply \emph{safe}) with respect to robot $j$ if, after executing the perturbed control \mbox{$\mathbf{\nu}_t\sim\mathcal{N}(\mathbf{u}_t, \Sigma)$} and transitioning to the next state $\mathbf{x}^i_{t+1}$, the probability that the inter-robot distance falls below $2r$ does not exceed a predefined threshold $\delta$. 

\textbf{The problem} now is to compute, at each time step, a control input $\mathbf{u}^i_t$ for every robot $i \in \mathcal{A}$ such that (i) it satisfies the control constraints given by \eqref{eq:control_bounds}; (ii) it ensures progress toward the assigned goal $\boldsymbol{\tau}_i$;
(iii) it remains probabilistically safe with respect to all observed neighboring robots.

\section{CoRL-MPPI: Cooperative RL-Guided MPPI}

\begin{figure*}[t!]
    \centering
    \includegraphics[width=0.99\linewidth]{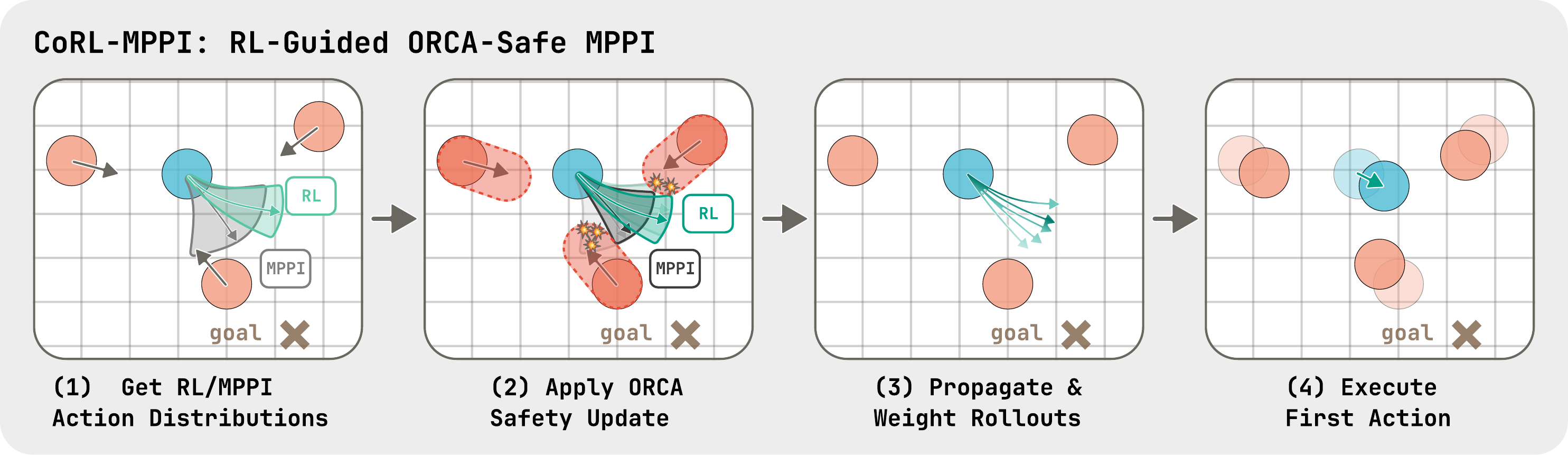}
    \caption{Overview of CoRL-MPPI. The robot observes its neighborhood and constructs two proposal distributions: an RL-guided one and a conventional MPPI one. ORCA-based constraints are used to safety-adjust both distributions in the control space. The resulting controls are sampled and propagated into rollouts, which are evaluated using the proposed cooperative cost function with symmetry breaking. The rollout costs are converted into MPPI importance weights and aggregated into a single control sequence. 
    Finally, its first action is executed.
    }
    \label{fig:corl_mppi_scheme}
\end{figure*}

The proposed method combines the safety-constrained MPPI framework with a cooperative reinforcement learning policy and a cooperative multi-agent cost function with symmetry breaking. At each planning step, candidate trajectories are sampled from two proposal distributions: an RL-guided branch and a conventional MPPI branch. The RL policy guides sampling toward coordinated behaviors, whereas the MPPI branch maintains exploration beyond the distribution induced by the learned policy, thereby reducing reliance on policy generalization. Safety is ensured through a constrained sampling procedure that provides formal safety guarantees and robustness to execution uncertainty. Furthermore, safety constraints are enforced over multiple prediction steps rather than only the immediate control action. The proposed multi-agent cost further promotes cooperative conflict resolution by encouraging greater separation between neighboring agents and introducing a consistent passing convention that breaks symmetric interactions. An overview of the resulting pipeline is shown in Fig.~\ref{fig:corl_mppi_scheme}, while its individual components are described below.

\subsection{MPPI and Safety-Constrained Sampling}
\label{sec:background_mppi}

The MPPI algorithm solves a finite-horizon stochastic optimal control problem in a receding-horizon manner. At each planning step, it optimizes a control sequence $u = (\mathbf{u}_0, \ldots, \mathbf{u}_{H-1})$ over a prediction horizon of length $H$. The objective consists of a terminal cost $\phi(\cdot)$, a state-dependent running cost $q(\cdot)$, and a control penalty term. A key advantage of MPPI is that both $\phi(\cdot)$ and $q(\cdot)$ can be defined in an arbitrary form, allowing domain-specific objectives and constraints to be incorporated directly into the trajectory evaluation process.

Given the current state $\mathbf{x}_0$ and a nominal control sequence $u^{init}$, MPPI generates $K$ candidate control sequences $\{u^k\}_{k=1}^{K}$ by sampling Gaussian perturbations around $u^{init}$ with sampling covariance $\Sigma^*$. Each sampled sequence induces a trajectory $x^k$ through the system dynamics and is assigned a trajectory cost $S(x^k,u^k)$, constructed from the terminal cost $\phi(\mathbf{x}^k_{H})$, running costs $q(\mathbf{x}^k_t)$, and a control penalty term~\cite{williams2017information}. The resulting costs are converted into importance weights used to update the nominal control sequence $u^{init}$. After executing the first control action, the optimized sequence is shifted forward and reused at the next planning step.

Since stochastic sampling may generate unsafe controls, we adopt the safety-constrained distribution shaping approach proposed in~\cite{dergachev2024model,dergachev2025decentralized}. The sampling distribution parameters are adjusted from $(u_t^{init}, \Sigma^*)$ to $(\hat u_t^{init}, \hat \Sigma^*)$ by solving a convex optimization problem that maximizes the probability of satisfying predefined safety constraints while remaining close to the nominal distribution. In this work, ORCA-based velocity constraints are mapped to linear inequalities over the sampled control variables~\cite{van2011reciprocal_n}. The resulting distribution guarantees that sampled controls satisfy all constraints with a prescribed probability.

\subsection{Cooperative RL-based Policy}

\label{sec:obs_space}
Each agent's observation is represented as a flat vector encoding its relative goal position and the relative positions of the $k$ nearest neighboring agents within the sensing radius. Specifically,
\begin{equation}
\mathbf{o}^i =
\left[
\frac{d_g^i}{R}, \frac{\alpha_g^i}{\pi},
\frac{d_1^i}{R}, \frac{\alpha_1^i}{\pi},
\ldots,
\frac{d_k^i}{R}, \frac{\alpha_k^i}{\pi}
\right],
\end{equation}
where $R$ is the sensing radius, $d_g^i$ and $\alpha_g^i$ are the distance and heading-relative angle from agent $i$ to its goal, and $d_j^i$ and $\alpha_j^i$ are the distance and heading-relative bearing to the $j$-th nearest neighbor. Neighbor slots are sorted by distance; neighbors outside the sensing radius and unused slots are filled with the sentinel value $1.0$. Each agent's action consists of continuous control variables bounded according to the action interface used during training.

We train the proposal policy using Independent PPO (IPPO)~\cite{yu2022surprising}, a decentralized multi-agent variant of PPO~\cite{schulman2017proximal}. In IPPO, each robot optimizes an actor-critic objective from its own local observations and rewards while treating other robots as part of the environment. As the robots are homogeneous, we use parameter sharing: a single policy is learned and then executed independently by every robot.

The policies were trained in the CAMAR environment~\cite{pshenitsyn2026camar} using the Sample Factory framework~\cite{petrenko2020sample}. Since IPPO is model-free, the proposal policy is trained separately for each action and dynamics interface used by CoRL-MPPI, while the method itself remains agnostic to the concrete dynamics model.

\label{sec:reward}
At every simulation step, the reward for agent $i$ combines goal reaching, progress toward the goal, collision avoidance, and personal-space preservation:
\begin{equation}
\label{eq:rl_reward}
\Re^i =
\Re^i_{\text{goal}} -
\Re^i_{\text{col}} +
\Re^i_{\text{dist}} -
\Re^i_{\text{space}} .
\end{equation}
The individual components are defined as:
\begin{equation}
\label{eq:rl_reward_terms}
\begin{cases}
\Re^i_{\text{goal}} & = w_g, \text{if } \| \mathbf{x}^i_{t+1} - \tau_i \| \leq r_{\tau}; \\
\Re^i_{\text{col}} & = w_c, \text{if } \exists j \in \mathcal{A}: \| \mathbf{x}_{t+1}^i - \mathbf{x}_{t+1}^j \| < r^i + r^j; \\
\Re^i_{\text{dist}} & = w_p \left(\| \mathbf{x}^i_{t} - \tau_i \| - \| \mathbf{x}^i_{t+1} - \tau_i \|\right); \\
\Re^i_{\text{space}} & = w_s \frac{PS^i-d_{\min}^i}{PS^i},  \text{if } d_{\min}^i \leq PS^i .
\end{cases}
\end{equation}
Here $d_{\min}^i = \min_{j \neq i} \| \mathbf{x}_{t+1}^i -\mathbf{x}_{t+1}^j \|$ is the distance to the nearest neighboring agent, and $PS^i$ is the personal-space radius. The selected numerical values for the reward weights and personal-space radius are reported in the experimental setup. The resulting policy is therefore optimized to produce cooperative local navigation behavior, while CoRL-MPPI retains safety through the constrained sampling procedure described next.

\subsection{CoRL-MPPI Planning Step}

\begin{figure}[hb!]
    \centering
    \includegraphics[width=1.0\linewidth]{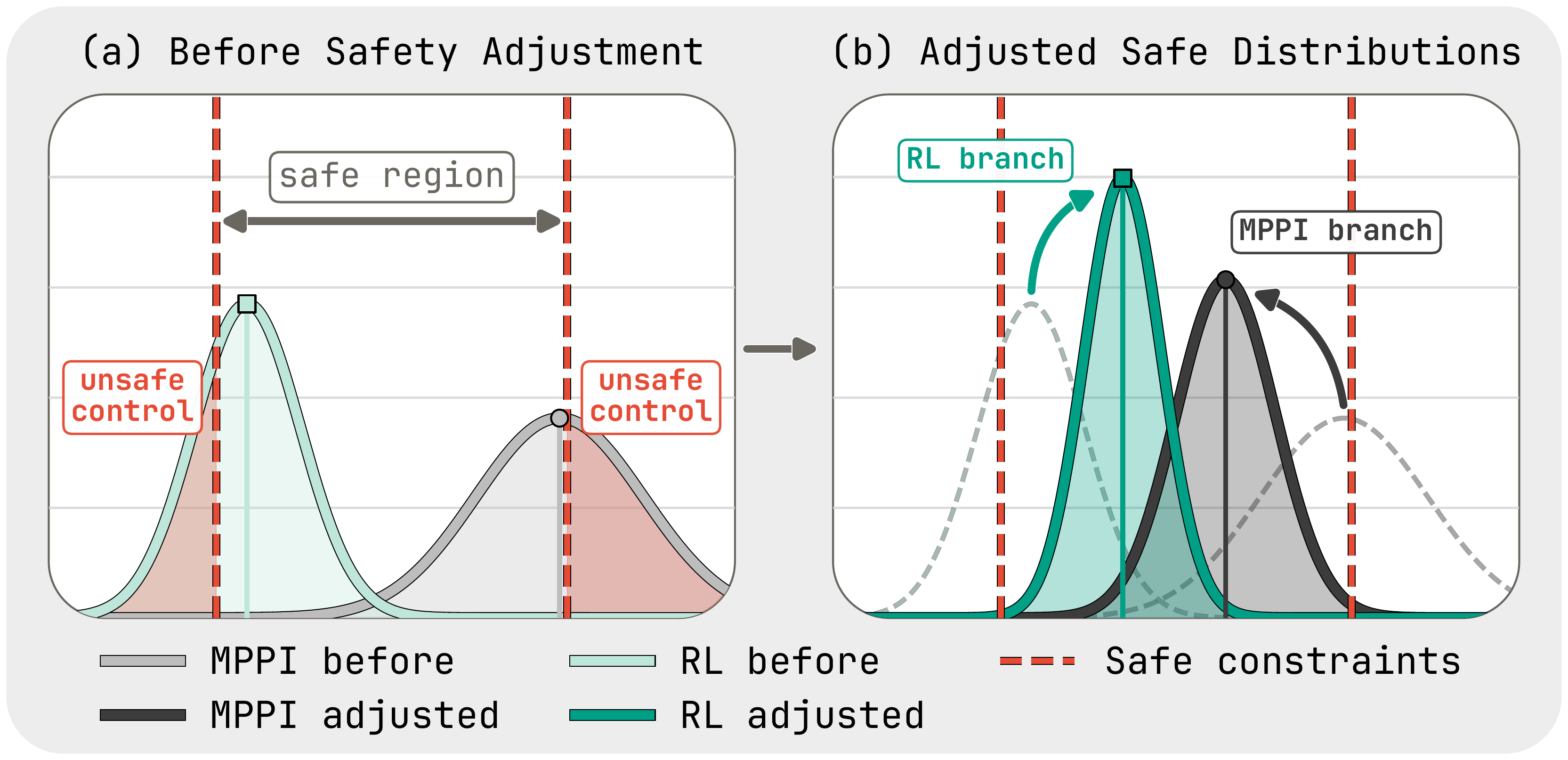}
    \caption{Control distributions before and after applying safety constraints. (a) Original MPPI and RL-guided distributions, with unsafe probability mass shown in red. (b) Distributions after the safety-constrained update, where the probability of sampling unsafe controls is reduced to the required confidence level. Dashed lines indicate safety bounds.
    }
    \label{fig:distr_vis}
\end{figure}

To account for future interactions, CoRL-MPPI first predicts the motion of neighboring agents over the planning horizon using a constant-velocity model. Two nominal proposal distributions are then maintained. The first corresponds to the standard MPPI initialization inherited from the previous planning iteration:
\begin{equation}
    u^{mppi} = u^{init}, \; \Sigma^{mppi} = \{\Sigma^{mppi}_t =\Sigma^*\}_{t=0}^{H-1}
\end{equation}
The RL-guided branch is generated using the pre-trained policy $\pi$. At each prediction step, the policy receives the local observation $\mathbf{o}^{i}_{t}$ recomputed from the predicted neighbor positions, the current predicted robot state, and the goal location. The policy outputs the parameters of a control distribution:
\begin{equation}
\mathbf{u}^{\pi}_{t}, \Sigma^{\pi}_{t} = \pi(\mathbf{x}^{\pi}_{t}, \mathbf{o}^{i}_{t}).
\end{equation}
Both branches are constructed recursively over the prediction horizon. At each step, a nominal control distribution is generated for the current predicted state. For the first $H_{safe}$ prediction steps, the safety-constrained distribution shaping procedure from~\cite{dergachev2025decentralized} is applied to obtain safe control distributions $\mathcal{N}(\hat{\mathbf{u}}^{{mppi}}_t,\hat{\Sigma}^{mppi}_t)$
and $\mathcal{N}(\hat{\mathbf{u}}^{\pi}_t,\hat{\Sigma}^{\pi}_t)$. The shaping procedure is shown in Figure~\ref{fig:distr_vis}. The corresponding mean control is then used to propagate the branch to the next predicted state, after which the process is repeated for the subsequent prediction step. Unlike MPPI-ORCA, which applies this procedure only to the first control action, CoRL-MPPI enforces safety constraints over multiple future steps.

After both nominal branches have been generated, the rollout budget is divided between them. A subset of control sequences is sampled around the RL-guided branch, while the remaining are sampled around the MPPI branch. Each sampled control sequence induces a trajectory that is evaluated using the proposed cost functions $\phi(\mathbf{x}_{H})$ and $q(\mathbf{x}_{t})$. The resulting trajectory costs are converted into MPPI importance weights, which are used to compute the optimized control sequence $u^*$. Finally, the first control action is executed, while the remaining controls are shifted forward and reused as the initialization for the next planning iteration.

\subsection{Cooperative Cost Function}

The sampled trajectories are evaluated using a cost function that combines goal-reaching, smoothness, collision avoidance, and multi-agent interaction terms:
\begin{equation}
\begin{gathered}
\phi(\mathbf{x}_{H}) = \phi_{goal}(\mathbf{x}_{H}),\\
\begin{aligned}
q(\mathbf{x}_t) &=
q_{goal}(\mathbf{x}_t)\;+\;q_{smooth}(\mathbf{x}_t)\;+\;q_{col}(\mathbf{x}_t)\;+\;\ldots\;\\
&\quad
+\;q_{mindist}(\mathbf{x}_t)
+\;q_{avdist}(\mathbf{x}_t)
+\;q_{side}(\mathbf{x}_t).
\end{aligned}
\end{gathered}
\label{eq:cost_terms}
\end{equation}

Most terms follow standard MPPI practice: $\phi_{goal}$ and $q_{goal}$ encourage progress toward the goal, while $q_{smooth}$ and $q_{col}$ penalize abrupt changes in the initial velocity direction and safety-margin violations. The nearest-neighbor penalty $q_{mindist}$ discourages close encounters with the closest visible agent, whereas the average-neighbor penalty $q_{avdist}$ accounts for separation from all visible neighbors. Although these terms are based on standard distance penalties, their combination was found to produce more stable spacing behavior than either component alone. 

The main extension lies in the new symmetry-breaking term $q_{side}$, which reduces deadlocks and oscillatory behavior. For neighboring agents, this term evaluates their predicted relative positions with respect to a fixed reference direction defined by the initial ego-goal axis at the current CoRL-MPPI update. A sampled trajectory is penalized when a neighbor lies on the undesired side of this reference axis, as determined by the sign of the 2D cross product between the relative neighbor position and the goal-directed reference vector.

\section{EXPERIMENTAL EVALUATION}

\subsection{Benchmarking in Numerical Simulation}

\begin{figure}[htb!]
    \centering
    \includegraphics[width=1.0\linewidth]{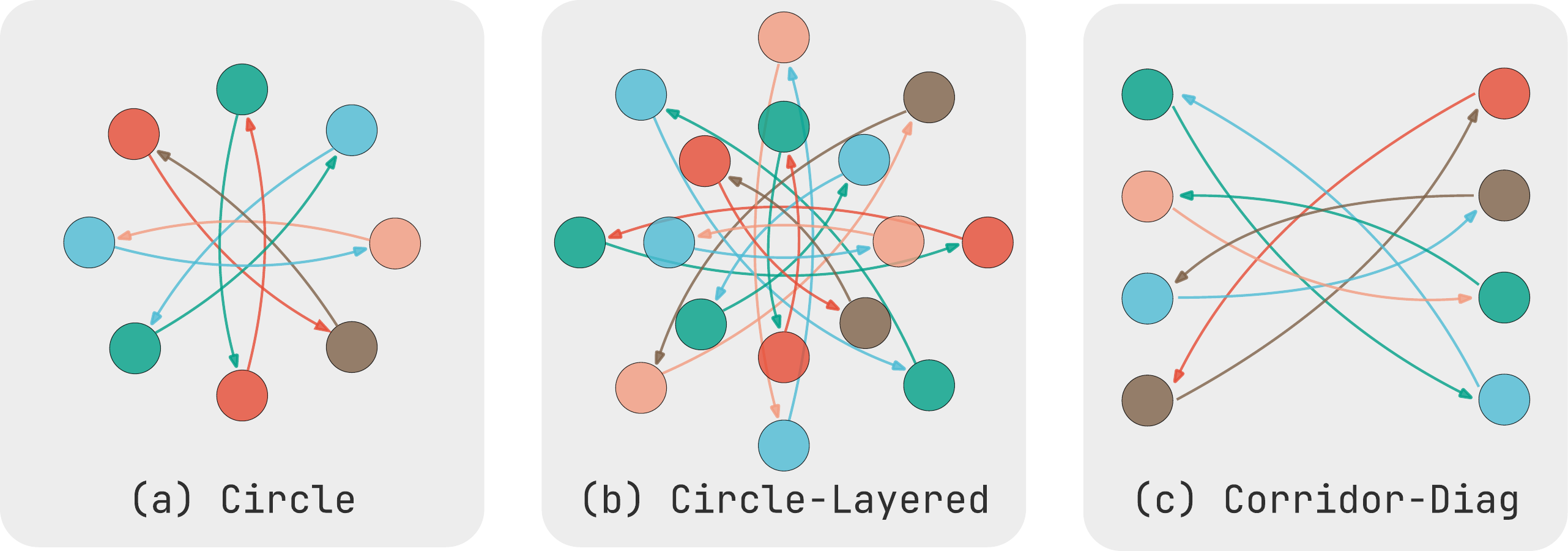}
    \caption{Illustrative visualization of the experimental scenarios. Scales and proportions are adjusted for clarity.}
    \label{fig:scenarios}
\end{figure}

\paragraph{\textbf{Experimental Setup}}

The experiments were conducted with two kinematic models: differential-drive and car-like robots~\cite{lavalle2006planning}. The differential-drive model is widely used in mobile robotics and enables direct comparison with a large body of prior multi-agent navigation methods. Cooperative collision avoidance for car-like robots is more challenging, as these robots cannot rotate in place, and only a limited number of prior works support such kinematics. 
To model imperfect control execution, zero-mean Gaussian noise $\varepsilon \sim \mathcal{N}(0, \Sigma)$ was added to the control inputs. The nominal controls were constrained according to~\eqref{eq:control_bounds}. All robots in the experiments shared identical physical and control parameters, with the corresponding values provided in Table~\ref{tab:simulation_parameters}.

\begin{table}[h!]
\centering
\footnotesize
\begin{tabular}{lcc}
\toprule
Parameter & Differential-Drive & Car-like \\
\midrule
Robot radius $r$ [m] & 0.30 & 0.34 \\
Wheelbase $L$ [m] & -- & 0.28 \\
$v$ limits [m/s] & $[-1.0,\,1.0]$ & $[-1.0,\,1.0]$ \\
$\omega\,,\,\phi$ limits [rad/s, rad] &
$[-2.0,\,2.0]$&
$[-\pi/4,\,\pi/4]$\\
Noise covariance $\Sigma$ &
$\operatorname{diag}(0.1,0.2)^2$ &
$\operatorname{diag}(0.1,0.05)^2$ \\
\midrule
\multicolumn{3}{c}{
Time step $\Delta t$: 0.1 s
\hfill
Observation radius: 15 m
} \\
\bottomrule
\end{tabular}
\caption{Agent and numerical simulation parameters used in the experiments.}
\label{tab:simulation_parameters}
\end{table}
Three different scenes were used for evaluation: \texttt{Circle}, \texttt{Circle-Layered}, and \texttt{Corridor-Diag} (Fig.~\ref{fig:scenarios}). The number of agents was varied within each scene, with the largest configuration reaching 96 agents. In \texttt{Circle}, agents were uniformly placed on a circle of diameter $14\,\text{m}$ and assigned goals on the opposite side. \texttt{Circle-Layered} follows the same interaction pattern but uses multiple concentric circles, each containing 8 agents. The innermost circle had a diameter of $6\,\text{m}$, each subsequent diameter was increased by $4\,\text{m}$, and the number of circles varied from 1 to 12. In \texttt{Corridor-Diag}, agents were placed in two opposite rows separated by $14\,\text{m}$, with $1\,\text{m}$ spacing within each row, and assigned diagonal goals in the opposite row. 

Notably, the circular scenes used in testing resemble the scenes used during RL policy training, while no scenarios resembling \texttt{Corridor-Diag} were seen during training. Thus, the latter scenes are \emph{out-of-distribution} and are used to assess the generalization of the proposed method.

\begin{figure*}[t]
\centering
    \begin{subfigure}[b]{\textwidth}
    \centering
    \includegraphics[width=1.0\textwidth]{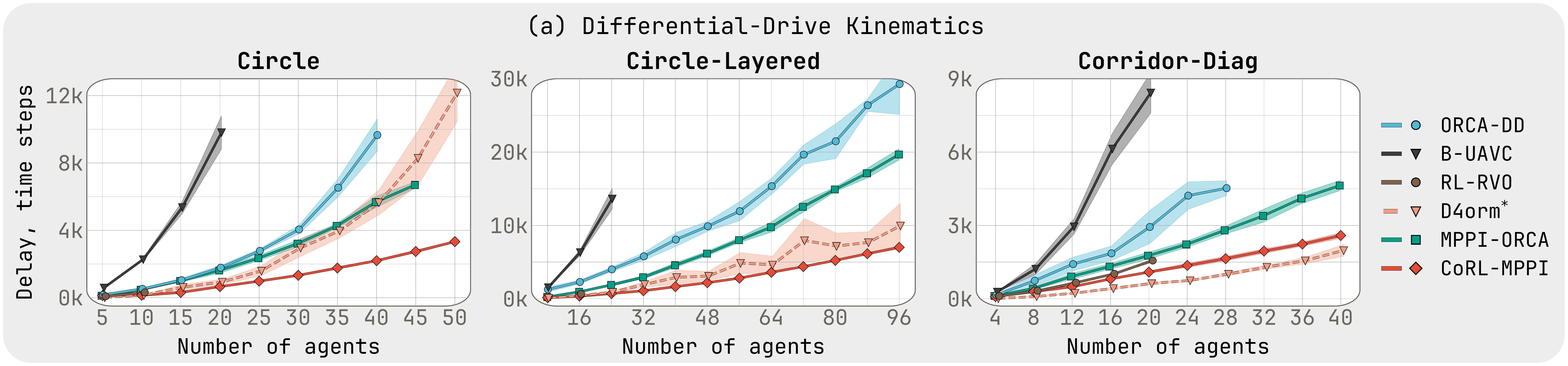}
    \end{subfigure}\\[0.2cm]
    
    \begin{subfigure}[b]{\textwidth}
    \centering
    \includegraphics[width=1.0\textwidth]{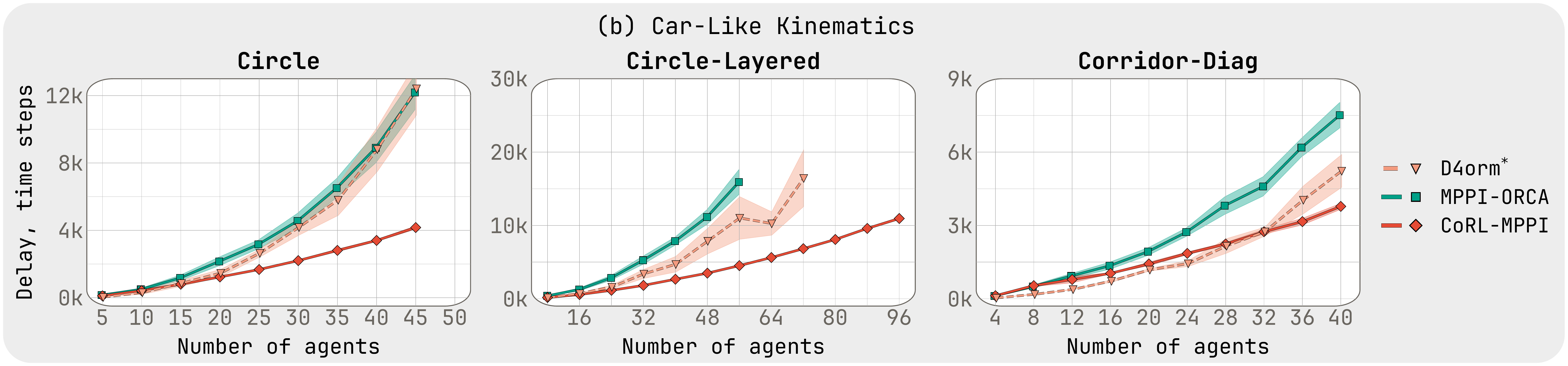}
    \end{subfigure}
    
\caption{Average \textbf{delay} of the evaluated algorithms. Lines indicate mean values and shaded regions show 95\% confidence intervals. Lower values are better. (*) \textsc{D4ORM} is a centralized planner, whereas all other methods are decentralized.}
\label{fig:delay_comparison}
\end{figure*}

\paragraph{\textbf{Algorithms and Implementation Details}}

Our method, CoRL-MPPI, was implemented in C++. Neural network inference was performed using ONNX Runtime.

 In all experiments we used a prediction horizon of 30 time steps ($3\,\text{seconds}$), with 1500 sampled rollouts per iteration. Among these, 50\% of the rollouts were sampled from the RL-guided distribution. Safety constraints were enforced over the first three steps of the prediction horizon. The weights of the cost-function components were tuned using Optuna.

% \footnote{\url{https://github.com/optuna/optuna}}
% \footnote{\url{https://github.com/microsoft/onnxruntime}}

 The RL-guided proposal policies used in these experiments were trained with IPPO and parameter sharing on 32-agent circle scenarios and 32-agent scenarios with random start-goal placement. Each policy was trained for $60M$ environment steps ($\approx 1.9B$ individual agent steps). For the reward in~\eqref{eq:rl_reward_terms}, the selected weights were $(w_g,w_c,w_p,w_s)=(1.75,4.6,3.7,4.0)$, and the personal-space radius was set to $PS^i=3r^i$. All RL training parameters were selected by grid search using validation success rate and collision frequency as the primary criteria.

For the differential-drive experiments, we compared CoRL-MPPI against decentralized and centralized baselines spanning classical, learning-based, and MPPI-based methods. Specifically, we compared against the decentralized methods \textsc{ORCA-DD}\cite{snape2010smooth}, \textsc{B-UAVC}\cite{zhu2022decentralized}, \textsc{RL-RVO}\cite{han2022reinforcement}, and \textsc{MPPI-ORCA}\cite{dergachev2025decentralized}, as well as the centralized planner \textsc{D4ORM}~\cite{zhang2025d4orm}. Since \textsc{D4ORM} assumes perfect execution of planned trajectories, it was executed in a receding-horizon manner and replanned at every simulation step using a planning horizon of 30 steps. For the car-like experiments, only \textsc{MPPI-ORCA} and \textsc{D4ORM} were considered, since the remaining baselines were specifically designed for differential-drive robots and would require substantial modifications or retraining.

\textsc{MPPI-ORCA} was configured using the same sampling parameters as CoRL-MPPI, but without the RL-guided sampling branch and new symmetry-breaking cost term. Its parameters were tuned using the same Optuna-based procedure.

To reduce deadlocks in highly symmetric scenarios, \textsc{ORCA-DD} and \textsc{B-UAVC} were modified by adding a small perturbation $(\varepsilon_x,\varepsilon_y)$, where $\varepsilon_x,\varepsilon_y \sim \mathcal{N}(0,0.3)$, to the goal direction vector. In addition, for \textsc{ORCA-DD}, \textsc{B-UAVC}, \textsc{MPPI-ORCA}, and \textsc{CoRL-MPPI}, the agent radius used during constraint computation was increased by $0.01\,\mathrm{m}$ relative to the physical robot radius to compensate for numerical errors.

\begin{table}[t]
\centering
\footnotesize
\setlength{\tabcolsep}{2.5pt}

\begin{tabular}{lcccccc}
\toprule
\footnotesize
&\multicolumn{2}{c}{\texttt{Circle}}
&\multicolumn{2}{c}{\texttt{Circle-L.}}
&\multicolumn{2}{c}{\texttt{Corridor-D.}} \\

\textbf{Algorithm} 
& \textbf{SR\,{\footnotesize$\uparrow$}} & \textbf{\%Col.\,{\footnotesize$\downarrow$}} 
& \textbf{SR\,{\footnotesize$\uparrow$}} & \textbf{\%Col.\,{\footnotesize$\downarrow$}} 
& \textbf{SR\,{\footnotesize$\uparrow$}} & \textbf{\%Col.\,{\footnotesize$\downarrow$}} \\
\midrule
\multicolumn{7}{c}{\texttt{Differential-Drive Kinematics}}\\
\midrule
\textsc{ORCA-DD} & 81\%	&13\% &80\% &0.8\% &66\% &1\%\\
\textsc{B-UAVC}  & 43\% &52\% &20.8\% &75.8\% &47\% &43\%\\
\textsc{RL-RVO} &19\% &81\% &5\% &95\% &49\% &51\%\\
\textsc{D4ORM$^{*}$} &\secondcell{99\%} &1\% &\secondcell{98.3\%} &1.7\% & \bestcell{100\%} & \bestcell{0\%} \\
\textsc{MPPI-ORCA} &92\% & \bestcell{0\%} &95\% &5\% &\bestcell{100\%} &\bestcell{0\%} \\
\textsc{CoRL-MPPI} (ours) &\bestcell{100\%} &\bestcell{0\%} &\bestcell{100\%} &\bestcell{0\%} &\bestcell{100\%} & \bestcell{0\%} \\

\midrule
\multicolumn{7}{c}{\texttt{Car-Like Kinematics}}\\
\midrule
\textsc{D4ORM$^{*}$} &\secondcell{90\%} &\bestcell{1\%} &72.5\% &25.8\% &97\% &\bestcell{0\%} \\
\textsc{MPPI-ORCA} &84\% &8\% &58.3\%	&40\% &\bestcell{100\%} &\bestcell{0\%} \\
\textsc{CoRL-MPPI} (ours)  &\bestcell{94\%} &3\% &\bestcell{100\%} &\bestcell{0\%} &\bestcell{100\%} &\bestcell{0\%} \\

\bottomrule
\end{tabular}

\caption{Success rate and percentage of runs terminated due to collisions. The arrows indicate preferred directions of improvement. Best results are highlighted in green and selected second-best success rates are highlighted in blue. (*) \textsc{D4ORM} is the only centralized planner, whereas all other methods are decentralized.}
\label{tab:sr_comparison}
\end{table}

\begin{table*}[th!]
\centering
\footnotesize
\setlength{\tabcolsep}{5pt}
\renewcommand{\arraystretch}{1.019}
% \begin{tabular}{lcccccccccccc}
\begin{tabular}{lrrrcrrrcrrrc}
\toprule
&\multicolumn{4}{c}{\texttt{Circle}}
&\multicolumn{4}{c}{\texttt{Circle-Layered}}
&\multicolumn{4}{c}{\texttt{Corridor-Diag}} \\
\textbf{Algorithm} 
& \textbf{SR\,{\small$\uparrow$}} & \textbf{\%Col.\,{\small$\downarrow$}} & \textbf{Delay\,{\small$\downarrow$}} & \textbf{$\Delta$\,{\small$\downarrow$}} 
& \textbf{SR\,{\small$\uparrow$}} & \textbf{\%Col.\,{\small$\downarrow$}} & \textbf{Delay\,{\small$\downarrow$}} & \textbf{$\Delta$\,{\small$\downarrow$}} 
& \textbf{SR\,{\small$\uparrow$}} & \textbf{\%Col.\,{\small$\downarrow$}} & \textbf{Delay\,{\small$\downarrow$}} & \textbf{$\Delta$\,{\small$\downarrow$}} \\
\cmidrule(r){1-1} \cmidrule(lr){2-5} \cmidrule(lr){6-9} \cmidrule(l){10-13}
\textsc{MPPI-ORCA} & 92\% & 0\% & \delayvalue{orcaCircleFull}{2820} & \deltaspace & 95\% & 5\% & \delayvalue{orcaLayerFull}{8464} & \deltaspace & 100\% & 0\% & \delayvalue{orcaCorridorFull}{2166} & \deltaspace\\
\textsc{MPPI-ORCA-LS} & 92\% & 0\% & \delayvalue{orcaCircleLs}{3596} & \deltaspace & 96.7\% & 3.3\% & \delayvalue{orcaLayerLs}{10960} & \deltaspace & 100\% & 0\% & \delayvalue{orcaCorridorLs}{2801} & \deltaspace\\

\midrule
\textsc{CoRL} & 37\% & 53\% & -- & & 71.7\% & 28.3\% & 4380 & & 7\% & 92\% & -- & \\
\midrule

\textsc{CoRL-MPPI} & 100\% & 0\% & \delayvalue{corlCircleFull}{1138} & \deltaspace & 100\% & 0\% & \delayvalue{corlLayerFull}{3199} & \deltaspace & 100\% & 0\% & \delayvalue{corlCorridorFull}{1256} & \deltaspace\\
\textsc{CoRL-MPPI-LS} & 100\% & 0\% & \delayvalue{corlCircleLs}{1193} & \deltaspace & 100\% & 0\% & \delayvalue{corlLayerLs}{3349} & \deltaspace & 100\% & 0\% & \delayvalue{corlCorridorLs}{1337} & \deltaspace\\

\bottomrule
\end{tabular}
\begin{tikzpicture}[remember picture, overlay]
\drawdelayarrow{orcaCircleFull}{orcaCircleLs}{\textbf{\textcolor{tblRed}{+27.5\%}}}
\drawdelayarrow{orcaLayerFull}{orcaLayerLs}{\textbf{\textcolor{tblRed}{+29.5\%}}}
\drawdelayarrow{orcaCorridorFull}{orcaCorridorLs}{\textbf{\textcolor{tblRed}{+29.3\%}}}
\drawdelayarrow{corlCircleFull}{corlCircleLs}{\textbf{\textcolor{tblGreen}{+4.8\%}}}
\drawdelayarrow{corlLayerFull}{corlLayerLs}{\textbf{\textcolor{tblGreen}{+4.7\%}}}
\drawdelayarrow{corlCorridorFull}{corlCorridorLs}{\textbf{\textcolor{tblGreen}{+6.4\%}}}
\end{tikzpicture}
\caption{Ablation study of CoRL-MPPI under differential-drive kinematics. The table reports success rate (SR), collision rate (\%Col.), and average delay for each scenario. The $\Delta$ annotations show the relative delay increase when reducing the number of sampled trajectories from 1500 to 250 for \textsc{CoRL-MPPI} and \textsc{MPPI-ORCA}. Delay values for the standalone RL policy are omitted in \texttt{Circle} and \texttt{Corridor-Diag} scenarios due to its low success rate.}
\label{tab:ablation}
\end{table*}
\begin{figure*}[th!]
    \centering
    \includegraphics[width=1.0\textwidth]{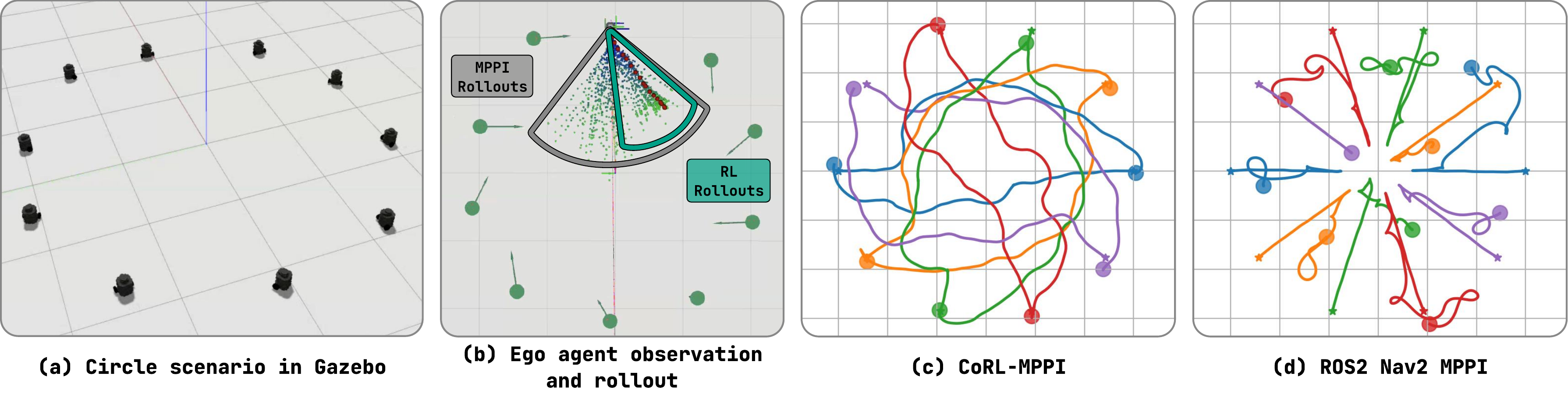}
    \caption{Simulation in Gazebo. (a) Ten TurtleBot3 robots arranged in the \texttt{Circle} scenario. (b) Local observation of a single robot and trajectory rollouts generated by CoRL-MPPI. The gray region shows MPPI-prior rollouts, while the blue region shows RL-guided proposal rollouts. (c)-(d) Trajectories produced by CoRL-MPPI and the built-in Nav2 MPPI Controller.}
    \label{fig:gazebo}
\end{figure*}

\paragraph{\textbf{Experimental Results}}

For each scene and agent count, 10 independent runs were performed. The following performance indicators were tracked. The \textbf{success rate} is the fraction of runs in which all agents reached their goals without collisions, within a tolerance of $0.3\,\text{m}$, before the limit of 1000 simulation steps, while not being required to remain at the goal after arrival. To obtain a more granular safety assessment, we also measured the percentage of runs terminated due to \textbf{collisions}. Finally, \textbf{delay} measures the additional travel time caused by multi-agent interactions compared to the straight-line motion at maximum speed.  

Table~\ref{tab:sr_comparison} reports the success rates and the proportion of runs terminated due to collisions. For differential-drive robots, \textsc{CoRL-MPPI} achieved a 100\% success rate and zero collisions in all scenarios, clearly outperforming all decentralized competitors. Among the decentralized baselines, \textsc{MPPI-ORCA} was the closest one, but produced failures and collisions in \texttt{Circle} and \texttt{Circle-Layered}. The centralized \textsc{D4ORM} planner also performed strongly, although it did not match the perfect success rate of our method. Other baselines, especially \textsc{B-UAVC} and \textsc{RL-RVO}, performed notably worse.

For car-like robots, \textsc{CoRL-MPPI} again performed best in terms of success rate, although rare failures and collisions occurred in \texttt{Circle}. Nevertheless, it substantially outperformed both \textsc{MPPI-ORCA} and the \textsc{D4ORM} planner, especially in the dense \texttt{Circle-Layered} scenario. 

Figure~\ref{fig:delay_comparison} presents the \textbf{delay} as a function of the number of agents. Each curve shows the mean delay computed over successful runs, while the shaded region denotes the corresponding 95\% confidence interval of the mean. A data point was included only if at least 5 out of 10 runs were successful. Furthermore, \textsc{RL-RVO} was omitted from the \texttt{Circle} and \texttt{Circle-Layered} plots due to very low success rate.

As seen in the figure, \textsc{CoRL-MPPI} achieved the lowest delay among all decentralized methods in all considered scenarios. In addition, it consistently exhibited the smallest variability across runs, resulting in extremely narrow confidence intervals that are often barely visible in the plots. This indicates not only high efficiency but also stable behavior across repeated executions. The only method that outperformed \textsc{CoRL-MPPI} was the centralized planner \textsc{D4ORM} in the \texttt{Corridor-Diag} scenario. However, for car-like robots the performance gap gradually decreased as the number of agents increased, and \textsc{CoRL-MPPI} eventually surpassed \textsc{D4ORM} in the most crowded settings. 

Overall, the results demonstrate that \textsc{CoRL-MPPI} provides the most robust performance among the evaluated decentralized methods, combining high success rates, low collision rates, and consistently low delay. Its advantage is especially pronounced in dense scenarios and under car-like kinematics, where cooperative collision avoidance becomes more challenging.

\paragraph{\textbf{Ablation}}

In addition to the main experiments, we conducted an ablation study of \textsc{CoRL-MPPI}. Specifically, we evaluated the standalone RL policy without MPPI, denoted as \textsc{CoRL}. We also evaluated a low-sampling variant, \textsc{CoRL-MPPI-LS}, where the number of sampled trajectories was reduced from 1500 to 250, a 6-fold reduction. For comparison, we also report \textsc{MPPI-ORCA} and its low-sampling variant, \textsc{MPPI-ORCA-LS}. The results are presented in Table~\ref{tab:ablation}.

The standalone RL policy achieved substantially lower success rates and significantly more collisions than the complete framework, especially in \texttt{Corridor-Diag}, indicating that it serves primarily as a source of informative sampling guidance rather than a complete navigation solution. In contrast, reducing the number of sampled trajectories from 1500 to 250 had only a minor effect on CoRL-MPPI, while noticeably degrading the performance of \textsc{MPPI-ORCA}. For example, in \texttt{Circle-Layered}, the normalized delay increased by only 4\% for CoRL-MPPI, compared to approximately 30\% for \textsc{MPPI-ORCA}. These results suggest that RL-guided sampling substantially improves the sample efficiency of the MPPI optimization process, allowing CoRL-MPPI to maintain stable performance even under a reduced computational budget.

\subsection{Experiments in Physics-Based Simulation}

To further evaluate the proposed approach, we validated it in Gazebo, a widely used physics-based robotics simulator. Specifically, we integrated \textsc{CoRL-MPPI} into the ROS2 Navigation Stack (Nav2)~\cite{macenski2020marathon2} as a custom local controller node. Global guidance was provided by a straight-line reference toward the goal. We compared against the built-in Nav2 MPPI Controller, which also uses MPPI for local collision avoidance. 

The experiments employed 10 differential-drive TurtleBot3 robots arranged in the \texttt{Circle} scenario. Each robot was controlled independently and received the ground-truth poses of neighboring robots through dedicated ROS topics at 8 ms intervals. A total of 10 runs were performed for each controller.

Figures~\ref{fig:gazebo}~(a)-(b) illustrate the Gazebo setup as well as the local observations of a single robot and the trajectory rollouts generated by CoRL-MPPI. The rollouts produced by the standard MPPI sampling process and the RL-guided proposal form two clearly distinguishable trajectory bundles. While the conventional MPPI samples explore a broad region of the control space, the RL-guided rollouts are concentrated around a promising maneuver proposed by the learned policy. Across all runs, our method consistently brought all robots to their assigned targets without collisions while preserving real-time operation with control updates above 10 Hz. By comparison, the built-in Nav2 MPPI Controller struggled to resolve multi-robot interactions (see Figure~\ref{fig:gazebo}~(c)-(d)).

\section{CONCLUSIONS}

In this work, we presented CoRL-MPPI, a hybrid framework that enhances Model Predictive Path Integral control with learned cooperative behavior for decentralized multi-robot collision avoidance. Our approach addresses a key limitation of vanilla MPPI, its reliance on uninformed random sampling, by using a pre-trained RL policy to bias the sampling distribution toward coordinated maneuvers. Extensive simulation experiments confirm that the proposed method significantly outperforms state-of-the-art decentralized baselines, including geometric, learning-based, and MPPI-based collision-avoidance methods. Promising directions include deploying the method on physical robot swarms to bridge the sim-to-real gap and investigating online policy adaptation for better generalization in diverse, evolving environments.

%                                   % the textheight of the last page by a suitable amount.
%                                   % This command does not take effect until the next page
%                                   % so it should come on the page before the last. Make
%                                   % sure that you do not shorten the textheight too much.

% %%%%%%%%%%%%%%%%%%%%%%%%%%%%%%%%%%%%%%%%%%%%%%%%%%%%%%%%%%%%%%%%%%%%%%%%%%%%%%%%

\bibliographystyle{IEEEtran}
\bibliography{IEEEabrv,bib}

@inproceedings{hansen2024td,
  title={Td-mpc2: Scalable, robust world models for continuous control},
  author={Hansen, Nick and Su, Hao and Wang, Xiaolong},
  booktitle={International Conference on Learning Representations},
  volume={2024},
  pages={47376--47405},
  year={2024}
}

@article{schulman2017proximal,
  title={Proximal policy optimization algorithms},
  author={Schulman, John and Wolski, Filip and Dhariwal, Prafulla and Radford, Alec and Klimov, Oleg},
  journal={arXiv preprint arXiv:1707.06347},
  year={2017}
}

@inproceedings{pshenitsyn2026camar,
  title={Camar: Continuous actions multi-agent routing},
  author={Pshenitsyn, Artem and Panov, Aleksandr and Skrynnik, Alexey},
  booktitle={Proceedings of the AAAI Conference on Artificial Intelligence},
  volume={40},
  number={35},
  pages={29651--29659},
  year={2026}
}

@inproceedings{petrenko2020sample,
  title={Sample factory: Egocentric 3d control from pixels at 100000 fps with asynchronous reinforcement learning},
  author={Petrenko, Aleksei and Huang, Zhehui and Kumar, Tushar and Sukhatme, Gaurav and Koltun, Vladlen},
  booktitle={International Conference on Machine Learning},
  pages={7652--7662},
  year={2020},
  organization={PMLR}
}

@article{han2022reinforcement,
  title={Reinforcement learned distributed multi-robot navigation with reciprocal velocity obstacle shaped rewards},
  author={Han, Ruihua and Chen, Shengduo and Wang, Shuaijun and Zhang, Zeqing and Gao, Rui and Hao, Qi and Pan, Jia},
  journal={IEEE Robotics and Automation Letters},
  volume={7},
  number={3},
  pages={5896--5903},
  year={2022},
  publisher={IEEE}
}

@article{bernstein2002complexity,
  title={The complexity of decentralized control of Markov decision processes},
  author={Bernstein, Daniel S and Givan, Robert and Immerman, Neil and Zilberstein, Shlomo},
  journal={Mathematics of operations research},
  volume={27},
  number={4},
  pages={819--840},
  year={2002},
  publisher={INFORMS}
}

@inproceedings{van2011reciprocal_n,
  title={Reciprocal n-body collision avoidance},
  author={Van Den Berg, Jur and Guy, Stephen J and Lin, Ming and Manocha, Dinesh},
  booktitle={Robotics Research: The 14th International Symposium ISRR},
  pages={3--19},
  year={2011},
  organization={Springer}
}

@article{zhou2017fast,
author = {Zhou, Dingjiang and Wang, Zijian and Bandyopadhyay, Saptarshi and Schwager, Mac},
journal = {IEEE Robotics and Automation Letters},
number = {2},
pages = {1047--1054},
title = {{Fast, on-line collision avoidance for dynamic vehicles using buffered voronoi cells}},
volume = {2},
year = {2017}
}

@inproceedings{snape2010smooth,
  title={Smooth and collision-free navigation for multiple robots under differential-drive constraints},
  author={Snape, Jamie and Van Den Berg, Jur and Guy, Stephen J and Manocha, Dinesh},
  booktitle={IEEE/RSJ International Conference on Intelligent Robots and Systems (IROS)},
  pages={4584--4589},
  year={2010}
}

@inproceedings{alonso2013optimal,
  title={Optimal reciprocal collision avoidance for multiple non-holonomic robots},
  author={Alonso-Mora, Javier and Breitenmoser, Andreas and Rufli, Martin and Beardsley, Paul and Siegwart, Roland},
  booktitle={Distributed autonomous robotic systems: The 10th international symposium},
  pages={203--216},
  year={2013}
}

@article{zhu2022decentralized,
  title={Decentralized probabilistic multi-robot collision avoidance using buffered uncertainty-aware Voronoi cells},
  author={Zhu, Hai and Brito, Bruno and Alonso-Mora, Javier},
  journal={Autonomous Robots},
  volume={46},
  number={2},
  pages={401--420},
  year={2022}
}

@inproceedings{long2018towards,
  title={Towards optimally decentralized multi-robot collision avoidance via deep reinforcement learning},
  author={Long, Pinxin and Fan, Tingxiang and Liao, Xinyi and Liu, Wenxi and Zhang, Hao and Pan, Jia},
  booktitle={IEEE International Conference on Robotics and Automation (ICRA)},
  pages={6252--6259},
  year={2018}
}

@inproceedings{williams2016aggressive,
  title={Aggressive driving with model predictive path integral control},
  author={Williams, Grady and Drews, Paul and Goldfain, Brian and Rehg, James M and Theodorou, Evangelos A},
  booktitle={IEEE International Conference on Robotics and Automation (ICRA)},
  pages={1433--1440},
  year={2016}
}

@inproceedings{williams2017information,
  title={Information theoretic MPC for model-based reinforcement learning},
  author={Williams, Grady and Wagener, Nolan and Goldfain, Brian and Drews, Paul and Rehg, James M and Boots, Byron and Theodorou, Evangelos A},
  booktitle={IEEE International Conference on Robotics and Automation (ICRA)},
  pages={1714--1721},
  year={2017}
}

@article{fan2020distributed,
  title={Distributed multi-robot collision avoidance via deep reinforcement learning for navigation in complex scenarios},
  author={Fan, Tingxiang and Long, Pinxin and Liu, Wenxi and Pan, Jia},
  journal={The International Journal of Robotics Research},
  volume={39},
  number={7},
  pages={856--892},
  year={2020},
  publisher={SAGE Publications Sage UK: London, England}
}

@article{gandhi2021robust,
  title={Robust model predictive path integral control: Analysis and performance guarantees},
  author={Gandhi, Manan S and Vlahov, Bogdan and Gibson, Jason and Williams, Grady and Theodorou, Evangelos A},
  journal={IEEE Robotics and Automation Letters},
  volume={6},
  number={2},
  pages={1423--1430},
  year={2021}
}

@inproceedings{tao2022control,
  title={Control barrier function augmentation in sampling-based control algorithm for sample efficiency},
  author={Tao, Chuyuan and Kim, Hunmin and Yoon, Hyungjin and Hovakimyan, Naira and Voulgaris, Petros},
  booktitle={2022 American Control Conference (ACC)},
  pages={3488--3493},
  year={2022}
}

@article{dergachev2024model,
  title = {Model Predictive Path Integral for Decentralized Multi-Agent Collision Avoidance},
  author = {Dergachev, Stepan and Yakovlev, Konstantin},
  year = {2024},
  journal = {PeerJ Computer Science},
  volume = {10},
  pages = {e2220},
}

@InProceedings{macenski2020marathon2,
author = {Macenski, Steven and Martin, Francisco and White, Ruffin and Ginés Clavero, Jonatan},
title = {The Marathon 2: A Navigation System},
booktitle = {IEEE/RSJ International Conference on Intelligent Robots and Systems (IROS)},
year = {2020}
}

@inproceedings{dergachev2025decentralized,
  title={Decentralized uncertainty-aware multi-agent collision avoidance with model predictive path integral},
  author={Dergachev, Stepan and Yakovlev, Konstantin},
  booktitle={IEEE/RSJ International Conference on Intelligent Robots and Systems (IROS)},
  pages={12456--12463},
  year={2025},
  organization={IEEE}
}

@inproceedings{yin2022trajectory,
  title={Trajectory distribution control for model predictive path integral control using covariance steering},
  author={Yin, Ji and Zhang, Zhiyuan and Theodorou, Evangelos and Tsiotras, Panagiotis},
  booktitle={IEEE International Conference on Robotics and Automation (ICRA)},
  pages={1478--1484},
  year={2022}
}

@article{qu2024rldriven,
  title = {{{RL-driven MPPI}}: {{Accelerating}} Online Control Laws Calculation with Offline Policy},
  author = {Qu, Yue and Chu, Hongqing and Gao, Shuhua and Guan, Jun and Yan, Haoqi and Xiao, Liming and Li, Shengbo Eben and Duan, Jingliang},
  date = {2024},
  year={2024},
  journal = {IEEE Transactions on Intelligent Vehicles},
  volume = {9},
  number = {2},
  pages = {3605--3616}
}

@inproceedings{chen2017decentralized,
  title = {Decentralized Non-Communicating Multiagent Collision Avoidance with Deep Reinforcement Learning},
  booktitle = {IEEE International Conference on Robotics and Automation (ICRA)},
  author = {Chen, Yu Fan and Liu, Miao and Everett, Michael and How, Jonathan P},
  date = {2017},
  year={2017},
  pages = {285--292}
}

@book{lavalle2006planning, 
    place={Cambridge}, 
    title={Planning Algorithms}, 
    publisher={Cambridge University Press}, 
    author={LaValle, Steven}, 
    year={2006}
}

@inproceedings{zhang2025d4orm,
  title={D4orm: Multi-Robot Trajectories with Dynamics-aware Diffusion Denoised Deformations},
  author={Zhang, Yuhao and Okumura, Keisuke and Woo, Heedo and Shankar, Ajay and Prorok, Amanda},
  booktitle={IEEE/RSJ International Conference on Intelligent Robots and Systems (IROS)},
  pages={14118--14123},
  year={2025},
  organization={IEEE}
}

@article{yu2022surprising,
  title={The surprising effectiveness of ppo in cooperative multi-agent games},
  author={Yu, Chao and Velu, Akash and Vinitsky, Eugene and Gao, Jiaxuan and Wang, Yu and Bayen, Alexandre and Wu, Yi},
  journal={Advances in neural information processing systems},
  volume={35},
  pages={24611--24624},
  year={2022}
}

\end{document}